\documentclass[runningheads]{llncs}
\usepackage{graphicx}
\usepackage{amssymb}
\usepackage{url}

\usepackage[a4paper, left=4.4cm, right=4.4cm, top=5.2cm, bottom=5.2cm]{geometry}

\urldef{\mails}\path|abhijit.mahalunkar@mydit.ie, john.d.kelleher@dit.ie|
\newcommand{\keywords}[1]{\par\addvspace\baselineskip
\noindent\keywordname\enspace\ignorespaces#1}

\begin{document}

\mainmatter

\title{Using Regular Languages to Explore the Representational Capacity of Recurrent Neural Architectures}
\titlerunning{Regular Languages and Recurrent Neural Architectures}
\author{Abhijit Mahalunkar\orcidID{0000-0001-5795-8728} \and \\ John D. Kelleher\orcidID{0000-0001-6462-3248}}
\authorrunning{A. Mahalunkar and J. D. Kelleher}
\institute{Dublin Institute of Technology\\Ireland\\
\mails\\}

\toctitle{Regular Languages}
\tocauthor{Neural Architecture}

\maketitle

\begin{abstract}
The presence of Long Distance Dependencies (LDDs) in sequential data poses significant challenges for computational models. Various recurrent neural architectures have been designed to mitigate this issue. In order to test these state-of-the-art architectures, there is growing need for rich benchmarking datasets. However, one of the drawbacks of existing datasets is the lack of experimental control with regards to the presence and/or degree of LDDs. This lack of control limits the analysis of model performance in relation to the specific challenge posed by LDDs. One way to address this is to use synthetic data having the properties of subregular languages. The degree of LDDs within the generated data can be controlled through the \emph{k} parameter, length of the generated strings, and by choosing appropriate \emph{forbidden strings}. In this paper, we explore the capacity of different RNN extensions to model LDDs, by evaluating these models on a sequence of SP\emph{k} synthesized datasets, where each subsequent dataset exhibits a longer degree of LDD. Even though SP\emph{k} are simple languages, the presence of LDDs does have significant impact on the performance of recurrent neural architectures, thus making them prime candidate in benchmarking tasks.

\end{abstract}

\section{Introduction}
A Recurrent Neural Network (RNN) is able to model temporal data efficiently \cite{elman_1990}. In theory, RNNs are capable of modeling infinitely long dependencies. A long distance dependency (LDD) describes a contingency (or interaction) between two (or more) elements in a sequence that are separated by an arbitrary number of positions. LDDs often occur in natural language, for example in English there is a requirement for subjects and verbs to agree, compare: ``\emph{The \textbf{dog} in that house \textbf{is} aggressive}'' with  ``\emph{The \textbf{dogs} in that house \textbf{are} aggressive}''. However, in practice successfully training an RNN to model LDDs is still extremely difficult, due in-part to exploding or vanishing gradients \cite{hochreiter_1991,bengio_1994}. There have been significant advances in this domain, and various architectures have been developed to address the issue of LDDs \cite{lstm_1997,graves_2014,salton_2017,merity_2016,chang_2017,zilly_2017,voronstov_2017,henaff_2016}. Indeed, the fact that a number of RNN extensions are specifically designed to address the problem of modeling LDDs is a testament to the fundamental importance of the challenge posed by LDDs.

In order to test the representational capacity of these models and aide in future development of new models, there is a growing need for large datasets which  manifest various degrees of LDDs. Various benchmarking datasets and tasks which exhibit such properties are currently being employed  \cite{ptb_1993,merity_2016,lecun_bottou_1998,lstm_1997}. However, using them provides no experimental control over the degree of LDD these datasets exhibit. Although, the copy and add task \cite{lstm_1997} does have control over this factor, the dataset generated via this scheme does not possess comparable complexity with datasets sampled from real world sequential processes.

Strictly \emph{k}-Piecewise (SP\emph{k}) languages, as studied by Rogers et al. \cite{rogers_2010}, are proper subclasses of piecewise testable languages \cite{simon_1975}. SP\emph{k} languages are natural and can express some of the kinds of LDDs found in natural languages \cite{fu_heinz_2011,heinz_rogers_2010}. In relation to research on LDDs, SP\emph{k} languages are particularly interesting because by controlling the parameter \emph{k} and the length of the strings, one can control the maximum LDD in the dataset, and by choosing appropriate \emph{forbidden strings}, it is possible to simulate a natural dataset exhibiting a certain degree of LDD. These properties make SP\emph{k} languages prime candidate for benchmarking tasks.

\textbf{Contribution:} This research used a finite-state implementation of an SP\emph{2} grammar to generate strings of varying length, from 2 to 500. SP\emph{2} is analogous to subject-verb agreement in English language, thus using this grammar generates LDDs of similar complexity, and controlling the length of the strings generated controls the maximum LDD span in the dataset. Appropriate \emph{forbidden strings} were chosen. State-of-the-art sequential data models were trained to predict the next character for every generated dataset. It was observed that as the length of the strings in the datasets increased the perplexity of the models increased. This is due in-part to the limitations of the representational capacity of these models. However, of the models tested it was observed that Recurrent Highway Networks display the lowest perplexity on character prediction task for large sequences exhibiting very high LDDs.

\section{Recurrent Neural Architectures for LDDs}\label{recurrentArch}
The focus of this paper is to experimentally evaluate the ability of modern RNN architectures to model LDDs by testing current state-of-the-art models on datasets of SP\emph{k} sequences which exhibit LDDs of varying lengths. For our experiments we chose the following architectures as the relevant representatives of RNNs: Long Short Term Memory \cite{lstm_1997}, Recurrent Highway Networks \cite{zilly_2017} and Orthogonal RNNs \cite{voronstov_2017}. This choice of networks was based on the fact that (a) each of these networks were specifically designed to address performance issue of the standard RNN while modeling LDD datasets, and (b) taken together the set of selected models provide coverage of the different approaches found in the literature to the problem of LDDs.

LSTMs were an early effort in addressing the vanishing gradient effect by introducing \emph{``constant error carousels"}, which enforced \emph{constant error flow through} thereby bridging minimal time lags in excess of 1000 discrete steps. Neural Turing Machines are memory augmented networks. They are composed of a network \emph{controller} and a memory bank. These components allowed the network to provide attention to different memory locations. Recurrent Highway Networks (RHNs) extended the LSTM architecture to allow step-to-step transition depths larger than one. Orthogonal RNNs (ORNNs) extend the standard RNN architecture by enforcing soft or hard orthogonality on the weight matrix.

\section{Benchmarking Datasets}
There is a relatively small number of datasets that are popular for testing the representational capacity of RNNs. Most of these datasets are known to exhibit LDDs, which is a necessary criteria for their selection as a benchmarking dataset. The \emph{Penn Treebank} \cite{ptb_1993} (PTB) is one of these datasets. It consists of over 4.5 million words of American English and was constructed by sampling English sentences from a range of sources.
The \emph{WikiText} language modeling dataset \cite{merity_2016} was released in 2016 and has become a popular choice for language modeling experiments. It is a collection of over 100 million tokens extracted from various Wikipedia articles. This dataset is much larger than the PTB, which is the primary reason that it is preferred to the PTB in recent works. Although, the PTB and WikiText differ in terms of the sources that the sentences they contain are sampled from, both dataset exclusively contain English language sentences. Hence both the datasets are constrained by English language grammar, and therefore will exhibit similar LDD characteristics. Moreover, it is unclear what these LDDs are because the data is sampled from a natural process (the English language) the LDD characteristics of which are not accurately estimated.

The difficulty of using naturally occurring datasets to investigate LDDs has been recognized and several synthetic benchmarks have been used to test the ability of RNNs to learn LDDs in sequential data. The copy and adding tasks, introduced in \cite{lstm_1997}, is one such example. The task entails remembering an input sequence followed by a string of blank inputs. The sequence is terminated using a delimiter after which the network must produce the input sequence, ignoring the string of blanks inputs that follow the original sequence \cite{voronstov_2017}. This task provides an experimenter with a great degree of control over the length of LDD in the dataset they synthesize in order to train and test their models.

Another method of testing models on simulated LDDs, is to train them to learn the MNIST image classes \cite{lecun_bottou_1998}. This is achieved by sequentially feeding all the 784 pixels of a MNIST image to the model under test and then training the network to classify MNIST image category. Every image is fed to the network pixel by pixel, starting from the top left pixel and finishing at the bottom right pixel. This simulates LDDs of length 784 as the network has to remember all the 784 pixels in order to classify the images.

Formal languages, have previously been used to train RNNs and investigate their inner workings. The \emph{Reber grammar} \cite{reber_1967} was used to train various first order RNNs \cite{casey_1996,smith_1989}. The Reber grammar was also used as a benchmarking dataset in the original work on LSTM models  \cite{lstm_1997}. Regular languages, studied by Tomita \cite{tomita_1982}, were used to train RNNs to learn grammatical structures of the string. A very recent example of research using formal languages to evaluate RNNs is Avcu et. al. \cite{avcu_2017}. The work presented in this paper falls within this tradition of analysis, however it extends the previous research on using formal languages by: (a) broadening the variety of LDDs within the generated datasets, (b) evaluating a broader variety of models, and (c) using language model perplexity as the evaluation metric.

\section{Formal Language Theory and Regular Languages}
Formal Language Theory (FLT) finds its use in various domains of science. Primarily developed to study the computational basis of human language, FLT is now being used to extensively analyze any rule-governed system \cite{chomsky_1956,chomsky_1959,fitch_2012}. Regular languages are the simplest grammars (type-3 grammars) within the Chomsky hierarchy which are driven by regular expressions. Subregular languages, e.g. Strictly \emph{k}-Piecewise or Strictly \emph{k}-Local, are subclasses of regular languages. These languages can be identified by mechanisms much less complicated than Finite-State Automata. Many aspects of human language such as local and non local dependencies are similar to subregular languages \cite{jager_2012}, and there are certain types of LDDs in human language which allow finite-state characterization \cite{heinz_rogers_2010}. These types of LDD can be modeled using Strictly \emph{k}-Piecewise languages. 

\subsection{Strictly Piecewise Languages}

In order to explain how we used SP\emph{k} languages to generate datasets appropriate to our experimental goals it is first necessary to present an explanation of these languages. Following \cite{avcu_2017,fu_heinz_2011,rogers_2010}, a language \emph{L} is described by a finite set of symbols, i.e. an alphabet, denoted by \( \Sigma \). The symbols are analogous to words or characters in English, music notes in music theory, genes in genomics, etc. A set \( \Sigma \)* is a \emph{free monoid}, a set of finite sequences of zero or more elements from \(\Sigma \). For example, for \(\Sigma \)=\{\emph{a,b,c}\}, its \( \Sigma \)* contains all concatenations of \emph{a, b,} and \emph{c}: \{\emph{\( \lambda \), a, ab, ba, cac, acbabc, ...}\}. The string of length zero is denoted by \( \lambda \). \emph{w\textsubscript{i}} is the \emph{i\textsuperscript{th}} word/string (\emph{w}) of \emph{L}. The length of a string \emph{u} is denoted \( \vert u \vert \). A stringset (or Formal Language) is a subset of \( \Sigma \)*.

If \emph{u} and \emph{v} are strings, \emph{uv} denotes their concatenation. For all \emph{u},\emph{v},\emph{w},\emph{x} \( \in \Sigma \)*, if \emph{x=uwv}, then \emph{w} is a \emph{substring} of \emph{x}. For example, \emph{bc} is a \emph{substring} of \emph{abcd}, as concatenating \emph{a,bc,d} yields \emph{abcd}. Similarly, a string \emph{v} is a \emph{subsequence} of string \emph{w} iff \emph{v}=\( \sigma \)\textsubscript{1}\( \sigma \)\textsubscript{2}...\( \sigma \)\textsubscript{n} and \emph{w} \( \in \Sigma \)*\(\sigma \)\textsubscript{1}\( \Sigma \)*\(\sigma \)\textsubscript{2}\( \Sigma \)*... \( \Sigma \)*\( \sigma \)\textsubscript{n}\( \Sigma \)*, where \( \sigma \in \Sigma\). For example, string \emph{bd} is a \emph{subsequence} of length \emph{k}=2 of \emph{abcd}, \emph{acd} is a subsequence of length \emph{k}=3 of the same string \emph{abcd}, but string \emph{db} is \emph{not a subsequence} of \emph{abcd}. A \emph{subsequence} of length \emph{k} is called a \emph{k-subsequence}. Let subseq\textsubscript{\emph{k}}(\emph{w}) denote the set of subsequences of \emph{w} up to length \emph{k}.

A Strictly Piecewise grammar can be defined as a set of permissible subsequences.  The grammar \emph{G} is simply all strings whose \emph{k}-long \emph{subsequences} are permissible according to \emph{G}. Consider a language \emph{L}, consisting of \( \Sigma \)=\emph{\{a,b,c,d}\}. The grammar, \emph{G\textsubscript{SP\emph{2}}}=\{\emph{aa, ac, ad, ba, bb, bc, bd, ca, cb, cc, cd, da, db, dc, dd}\} are comprised of these permissible \emph{subsequences} of length \emph{k}=2. Note, however, that although \emph{\{ab}\} is a logically possible subsequence of length \emph{k}, it is not in the grammar. Subsequences which are not in the grammar are called \emph{forbidden strings}. The string \emph{u}=[\emph{bbcbdd}], where \( \vert \)\emph{u}\( \vert \)=6 belongs to \emph{G\textsubscript{SP\emph{2}}}, because it is composed of subsequences that are in that grammar. Similarly, the string \emph{v}=[\emph{bbdbbbcbddaa}], where \( \vert \)\emph{v}\( \vert \)=12 belongs to \emph{G\textsubscript{SP2}}. However, the string \emph{w}=[\emph{bbabbbcbdd}] does not because \emph{\{ab}\} is a forbidden subsequence as it is not part of the grammar. This condition applies for any string \emph{x} for \( \vert \)\emph{x}\( \vert \in \mathbb{Z} \). One can also define an SP grammar for k=3 and k=4 for \( \Sigma \)=\{\emph{a,b}\} as G\textsubscript{SP\emph{3}} and G\textsubscript{SP\emph{4}} respectively. For example, G\textsubscript{SP\emph{3}}=\{\emph{aaa, aab, abb, baa, bab, bba, bbb}\}, with \{\emph{aba}\} as \emph{forbidden string}. A string [\emph{aaaaaaab}] of length 8 is a valid \emph{G\textsubscript{SP3}} string and [\emph{aaaaabaa}] is invalid. Thus, an appropriate grammar reflecting the dataset one intends to simulate can be designed by selecting appropriate permissible strings in the grammar. For the specific language, \emph{forbidden strings} can be computed\footnote{Refer section \emph{5.2. Finding the shortest forbidden subsequences} in \cite{fu_heinz_2011} for method to compute \emph{forbidden sequences} for a particular SP language.}. Note, to define an SP\emph{k} grammar it is necessary to specify at least one \emph{forbidden string}.

\begin{figure}
\centering
\includegraphics[scale=0.375]{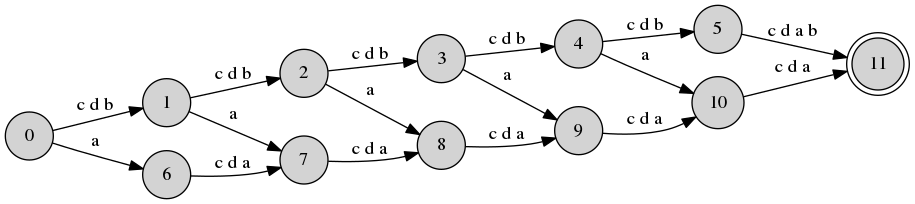}
\caption{The automaton for \emph{G\textsubscript{SP2}} (\emph{k}=2) which generates strings of length=6}
\label{fig:fsa1}
\end{figure}

Figure~\ref{fig:fsa1} illustrates the finite-state diagram of a \emph{G\textsubscript{SP2}} for strings of length 6 with \emph{forbidden string} \{\emph{ab}\}. Traversing a path from state 1 until state 11 will generate valid \emph{G\textsubscript{SP2}} strings of length 6, e.g. \{\emph{accdda, caaaaa}\}. It can also be noted that there is no path which generates a string which has an \{\emph{ab}\} subsequence e.g. \{\emph{abcccc}\} does not exist. Using the above described methodology, of choosing strings of appropriate length, one can simulate appropriate LDDs in a dataset. One can also control the number of dependent elements by choosing an appropriate \emph{k}. \emph{Forbidden strings} allow for elimination of certain combinations in generated datasets, which can be useful when one is trying to simulate real world datasets.

\section{Experiment}
In this experiment, we generate 4 datasets of SP\emph{2} language. For each dataset we train an LSTM, an  ORNN, and a RHN, and evaluate and compare the performance of the models.

\subsection{Generating SP\emph{2} dataset}
For our experiment, \( \Sigma \)=\{\emph{a,b,c,d}\} was selected. \emph{Forbidden strings} for this language were selected as \{\emph{ab,bc}\}. In order to introduce various degrees of LDDs, strings with lengths \emph{l} were generated in random order, where \( 2\leq \emph{l} \leq 500 \). For every \emph{l}, the number of strings per \emph{l} is \emph{n\textsubscript{l}}. For this experiment, \emph{n\textsubscript{l}} \( \leq \) 1,000,000. This allowed for uniform distribution of strings of all lengths. These strings were grouped in 4 datasets as described in Table~\ref{tab1}. Within each dataset, strings were randomly ordered to avoid biased gradients. For training the neural networks, a subset of these generated datasets were used due to the size of each dataset.

\begin{table}
\centering
\caption{Datasets of SP\emph{2} language}\label{tab1}
\begin{tabular}{|c|c|c|c|c|c|}
\hline
\ \ \ Dataset\ \ \  & \ \ \ Min Length\ \ \  & \ \ \ Max Length\ \ \ & \ \ \ Max LDDs\ \ \ & \ \ \ Original\ \ \ & \ \ \ Sample\ \ \ \\
\hline
Dataset 1 & 2 & 20 & 20 & 15 MB & 15 MB\\
Dataset 2 & 21 & 100 & 100 & 470 MB & 50 MB\\
Dataset 3 & 101 & 200 & 200 & 1.5 GB & 100 MB\\
Dataset 4 & 201 & 500 & 500 & 9.9 GB & 200 MB\\
\hline
\end{tabular}
\end{table}

The strings were generated using the tool \emph{foma} \cite{hulden_2009}. A post processing \emph{python} script was developed to select the small sample from the original datasets 1, 2, 3 and 4 as described in Table~\ref{tab1}. Every dataset is made up of strings of varying \emph{l}. The \emph{python script} was also used to randomize the order of strings (as per the length), so as not to bias the models\footnote{Source code available at \url{https://github.com/silentknight/ICANN2018}}.

\subsection{Training Task}
All the networks were trained on a character prediction task. For each network type (LSTM, ORNN, RHN) a network was trained on each of the 4 SP\emph{2} datasets, and also on a standard dataset of English language. The English language datasets were included in the experiments to provide a comparison for model performance when the vocabulary and type of data was varied. For the LSTM and ORNN the PTB was used as the standard English language dataset, and for the RHN the Text8 dataset was used. Note, that the experimental task was kept constant across all datasets, so although the PTB and Text8 datasets are often used as part of a word-prediction task, in these experiments the networks were trained and evaluated on character prediction on the PTB and Text8 datasets. For SP\emph{k} languages, the generated datasets were split into training (60\%), validation (20\%) and test (20\%) sets. The LSTM\footnote{LSTM source \url{https://github.com/tensorflow/models/blob/master/tutorials/rnn/ptb/ptb_word_lm.py}} with dropout models were trained as advised in \cite{zaremba_2015}; the ORNN\footnote{ORNN Source \url{https://github.com/veugene/spectre_release}} models were trained as recommended in \cite{voronstov_2017}; and, the RHN\footnote{RHN source \url{https://github.com/julian121266/RecurrentHighwayNetworks}} models were trained following \cite{zilly_2017}.

\begin{figure}
\centering
\begin{tabular}{cc}
\includegraphics[scale=0.35]{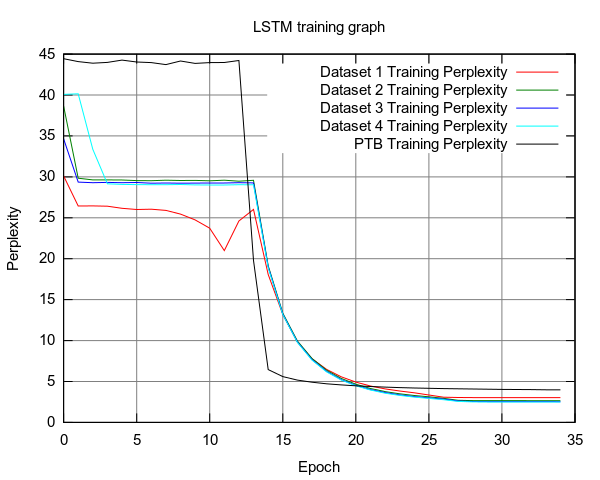} & \includegraphics[scale=0.35]{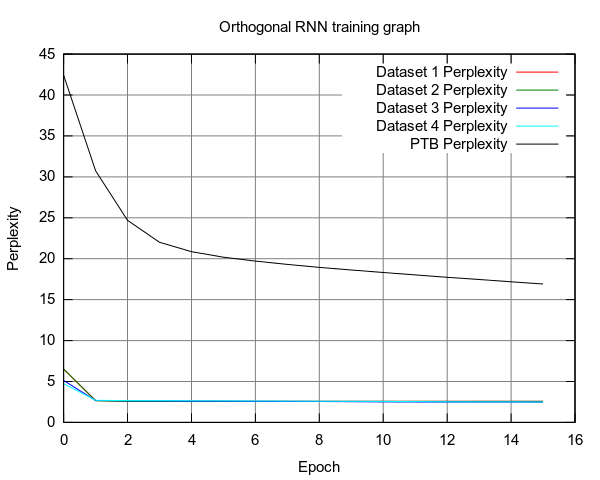} \\
(a) LSTM Network & (b) Orthogonal RNN \\
\end{tabular}
\includegraphics[scale=0.35]{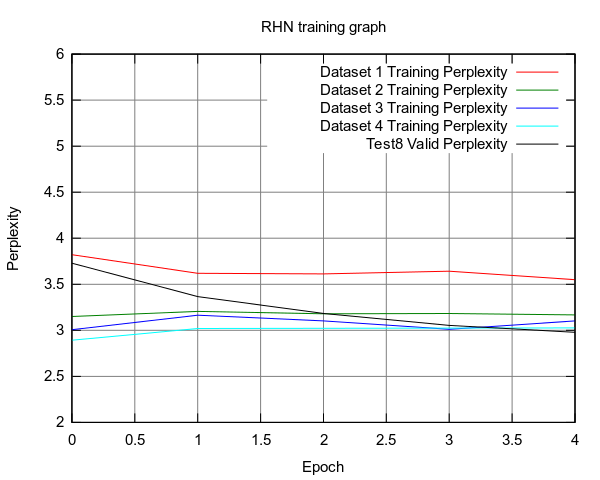} \\
(c) Recurrent Highway Network
\caption{Perplexity vs training epoch for recurrent neural architectures.}\label{performance}
\end{figure}

The performance of all the three network types was measured by computing the perplexity of the network after each epoch. The performance curve for the LSTM model is plotted in Figure~\ref{performance}.a., the performance of ORNN model is plotted in Figure~\ref{performance}.b., and the performance curve of RHN is plotted in Figure~\ref{performance}.c.

\section{Analysis}
In Figure \ref{performance}, we visualize the impact of increasing LDDs while training all the three architectures. Our results show that during the initial phase of the training, the LSTM network displayed perplexity directly proportional to the degree of LDDs present in the dataset. It is seen that dataset 4 (LDD order of around 500) presents higher perplexity as compared to the other datasets. However, every dataset eventually exhibits lower perplexity after epoch 20. When compared with the PTB task, one can observe lower perplexity by LSTM network in modeling datasets 1, 2, 3 and 4 during the initial phase of training.  This is due in-part to the small vocabulary size in the SP\emph{2} datasets (\( \Sigma \)=\{\emph{a,b,c,d}\}). A small vocabulary size tends to lower entropy in a sequence. The PTB has much larger vocabulary thus increasing the entropy and eventually increasing perplexity. Selection of more \emph{forbidden strings} leads to much richer grammar. SP\emph{k} languages generated for this experiment contained only 2 \emph{forbidden strings}, this led to generation of less rich grammar as compared to the PTB (English grammar). However, one can observe that the LSTM model learns the PTB much faster than SP\emph{2} languages (the graph drops earlier). This can be directly attributed to the presence of longer LDDs in the SP\emph{2} datasets.

Orthogonal RNNs enforce soft orthogonality to address vanishing gradient problem. When compared with LSTM network training of the PTB, it is observed that the perplexity of both architectures is very similar during the initial training phase, but ORNNs performance does not improve with more training as compared to LSTM. The impact of vocabulary size is also evident in this case (the perplexity for PTB is much higher than for the SP\emph{2} datasets). However, it can be seen that ORNNs trained with datasets 1 and 2 present higher perplexity as compared to datasets 3 and 4 (longer LDDs) suggesting that ORNN models overfit datasets 1 and 2 and are able to generalize to datasets 3 and 4. This could be attributed to orthogonal weight initializations which makes learning longer dependencies easier.

Focusing on the graph for the Recurrent Highway Networks it can be observed that the model tended to exhibit lower perplexity on SP\emph{2} datasets with higher degrees of LDDs. This could be attributed to the architecture of the network. Due to increased depth in recurrent transitions in these networks, it was possible for the model to achieve good performance on datasets with long LDDs. However, on datasets with lower degrees of LDDs these models tend to overfit and, thus, exhibit higher perplexity. Furthermore, comparing the RHN graph on the Text8 dataset with the LSTM and ORNN graphs on the PTB it is apparent that RHNs are better at handling larger vocabularies: the RHN graph for Text8 is lower than the LSTM and ORNN graphs on the PTB.

\section{Conclusion}
In this paper, we used SP\emph{k} languages to generate benchmarking datasets for LDDs. We trained various RNNs with the generated datasets and analyzed their performance. The analysis revealed that SP\emph{k} languages are able to generate datasets with varying degree of LDDs. Consequently, using SP\emph{k} languages gives experimental control over the generation of rich datasets by controlling the \emph{k}, the length of the strings, the vocabulary of the generated language, and by choosing appropriate \emph{forbidden strings}. The analysis also revealed that RHNs have a much better capability (as compared with LSTMs and ORNNs) to model LDDs.

\subsubsection{Acknowledgements}
This research was partly supported by the ADAPT Research Centre, funded under the SFI Research Centres Programme (Grant 13/ RC/2106) and is co-funded under the European Regional Development Funds. The research was also supported by an IBM Shared University Research Award. We also, gratefully, acknowledge the support of NVIDIA Corporation with the donation of the Titan Xp GPU under NVIDIA GPU Grant used for this research.


\begin{thebibliography}{4}

\bibitem{elman_1990}
Elman, J. L.: Finding Structure in Time. Cognitive Science \textbf{14}, pp. 179-211 (1990).

\bibitem{hochreiter_1991}
Hochreiter. S.: Untersuchungen zu dynamischen neuronalen Netzen. Diploma thesis, TU Munich (1991).

\bibitem{bengio_1994}
Yoshua, B., Simard, P., Frasconi, P.: Learning Long-Term Dependencies with Gradient Descent is Difficult. IEEE Transactions on Neural Networks \textbf{5}(2), pp. 157-166 (1994).

\bibitem{lstm_1997}
Hochreiter, S., Schmidhuber, J.: Long short-term memory. Neural Computation, \textbf{9}(8), pp. 1735-1780, (1997).

\bibitem{graves_2014}
Graves, A., Wayne, G., Danihelka, I.: Neural Turing Machines. CoRR (2014).

\bibitem{salton_2017}
Salton, G. D., Ross, R. J., Kelleher, J. D.: Attentive Language Models. In: Proceedings of the The 8th International Joint Conference on Natural Language Processing, pp. 441-450 (2017).

\bibitem{merity_2016}
Merity, S., Xiong, C., Bradbury, J., Socher, R.: Pointer Sentinel Mixture Models. In: ICLR 2016, (2016).

\bibitem{chang_2017}
Chang, S. et al.: Dilated Recurrent Neural Networks. In: Editors Guyon, I., Luxburg, U. V., Bengio, S., Wallach, H., Fergus, R., Vishwanathan, S., Garnett, R. (eds.), Advances in Neural Information Processing Systems 30, pp. 77-87. Curran Associates, Inc. (2017).

\bibitem{zilly_2017}
Zilly, J. G., Srivastava, R. K., Koutník, J., Schmidhuber, J.: Recurrent Highway Networks. In: Proceedings of the 34th International Conference on Machine Learning, Sydney, Australia, PMLR \textbf{70} (2017).

\bibitem{voronstov_2017}
Vorontsov, E., Trabelsi, C., Kadoury, S., Pal, C.: On orthogonality and learning recurrent networks with long term dependencies. In: Proceeding of ICML 2017 (2017).

\bibitem{henaff_2016}
Henaff, M., Szlam, A., LeCun, Y.: Recurrent Orthogonal Networks and Long-Memory Tasks. In: Proceedings of The 33rd International Conference on Machine Learning, in PMLR \textbf{48}, pp. 2034-2042 (2016).

\bibitem{ptb_1993}
Marcus, M. P., Marcinkiewicz, M. A., Santorini, B.: Building a large annotated corpus of english: The penn treebank. Computational Linguistics \textbf{19}(2), pp. 313-330 (1993). ISSN 0891-2017.

\bibitem{lecun_bottou_1998}
LeCun, Y., Bottou, L., Bengio, Y., Haffner,
P.: Gradient-based learning applied to document recognition. In: Proceedings of the IEEE \textbf{86}(11), pp. 2278-2324 (1998).

\bibitem{rogers_2010}
Rogers, J., Heinz, J., Bailey, G., Edlefsen, M., Visscher, M., Wellcome, D., Wibel, S.: On Languages Piecewise Testable in the Strict Sense. In: Ebert C., Jäger G., Michaelis J. (eds.) The Mathematics of Language. Lecture Notes in Computer Science, \textbf{6149}, Springer, Berlin, Heidelberg (2010).

\bibitem{simon_1975}
Simon, I.: Piecewise testable events. In: Automata Theory and Formal Languages, pp. 214-222 (1975).

\bibitem{fu_heinz_2011}
Fu, J., Heinz, J., Tanner, H.G.: An Algebraic Characterization of Strictly Piecewise Languages. In: Ogihara M., Tarui J. (eds.) Theory and Applications of Models of Computation. TAMC 2011. Lecture Notes in Computer Science, vol. 6648, Springer, Berlin, Heidelberg (2011).

\bibitem{avcu_2017}
Avcu, E., Shibata, C., Heinz, J.: Subregular Complexity and Deep Learning. In: Proceedings of the Conference on Logic and Machine Learning in Natural Language (LaML 2017), \textbf{1}, pp. 20-33 (2017).

\bibitem{heinz_rogers_2010}
Heinz. J., Rogers, J.: Estimating Strictly Piecewise Distributions. In: Proceedings of the 48th Annual Meeting of the Association for Computational Linguistics, pp. 886-896 (2010).

\bibitem{reber_1967}
Reber, A. S.: Implicit learning of artificial grammars. Journal of Verbal Learning and Verbal Behavior \textbf{6}(6), pp.855-863 (1967).

\bibitem{tomita_1982}
Tomita, M.: Learning of construction of finite automata from examples using hill-climbing. In: Proceedings of Fourth International Cognitive Science Conference, pp. 105-108 (1982).

\bibitem{casey_1996}
Casey, M.: The dynamics of discrete-time computation, with application to recurrent neural networks and finite statemachine extraction. Neural computation \textbf{8}(6), pp. 1135-78 (1996).

\bibitem{smith_1989}
Smith, A. W., Zipser, D.: Encoding sequential structure: experience with the real-time recurrent learning algorithm. In: Proceedings of IJCNN, vol. I, pp. 645-648 (1989).

\bibitem{chomsky_1956}
Chomsky, N.: Three models for the description of language. IRE Transactions on Information Theory \textbf{2}, pp 113-124 (1956).

\bibitem{chomsky_1959}
Chomsky, N.: On certain formal properties of grammars. Information Control \textbf{2}, pp 137-167 (1959).

\bibitem{fitch_2012}
Fitch, W. T., Friederici, A. D.: Artificial grammar learning meets formal language theory: an overview. Philosophical Transactions of Royal Society B: Biological Sciences \textbf{367}(1598), pp. 1933-1955 (2012).

\bibitem{jager_2012}
Jager, G., Rogers, J.: Formal language theory: refining the Chomsky hierarchy. Philosophical Transactions of Royal Society B: Biological Sciences \textbf{367}(1598), pp. 1956-1970 (2012).

\bibitem{ebeling_1994}
Ebeling, W., Poeschel, T.: Entropy and Long range correlations in literary English. Europhysics Letters \textbf{26}(2), pp. 241-246 (1994).

\bibitem{hulden_2009}
Hulden, M.: Foma: a finite-state compiler and library. Proceedings of the 12th Conference of the European Chapter of the Association for Computational Linguistics, pp. 29-32 (2009).

\bibitem{zaremba_2015}
Zaremba, W., Sutskever, I., Vinyals, O.: Recurrent Neural Network Regularization. In: Proceedings of ICRL (2015). 

\end{thebibliography}
\end{document}